\documentclass[letterpaper]{article} 
\usepackage{aaai2026}  
\usepackage{times}  
\usepackage{helvet}  
\usepackage{courier}  
\usepackage[hyphens]{url}  
\usepackage{graphicx} 
\usepackage{natbib}  
\usepackage{caption} 
\usepackage{algorithm}
\usepackage{algorithmic}
\usepackage[hidelinks]{hyperref}

\usepackage[utf8]{inputenc} 
\usepackage{amsmath}        
\usepackage{amssymb}
\usepackage{colortbl}
\usepackage{xcolor}
\usepackage{etoolbox}
\AtEndPreamble{
    \usepackage[capitalize]{cleveref}
    \crefname{section}{Sec.}{Secs.}
    \Crefname{section}{Section}{Sections}
    \crefname{table}{Tab.}{Tabs.}
    \Crefname{table}{Table}{Tables}
}
\usepackage{afterpage}

\usepackage{newfloat}
\usepackage{listings}
\DeclareCaptionStyle{ruled}{labelfont=normalfont,labelsep=colon,strut=off} 
\floatstyle{ruled}
\newfloat{listing}{tb}{lst}{}
\floatname{listing}{Listing}
\title{GUIDE: Gaussian Unified Instance Detection \\
for Enhanced Obstacle Perception in Autonomous Driving}
\author{
    Chunyong Hu\textsuperscript{\rm 1}\equalcontrib,
    Qi Luo\textsuperscript{\rm 1}\equalcontrib,
    Jianyun Xu\textsuperscript{\rm 1}\thanks{Project leader.},
    Song Wang\textsuperscript{\rm 1,2},
    Qiang Li\textsuperscript{\rm 1},
    Sheng Yang\textsuperscript{\rm 1}\thanks{Corresponding author.}
}
\affiliations{
    \textsuperscript{\rm 1}Unmanned Vehicle Dept., CaiNiao Inc., Alibaba Group \:
    \textsuperscript{\rm 2}Zhejiang University\\
}

\usepackage{bibentry}

\begin{document}
\maketitle

\afterpage{%
\begin{figure*}[t]
  \centering
  \includegraphics[width=0.95\textwidth]{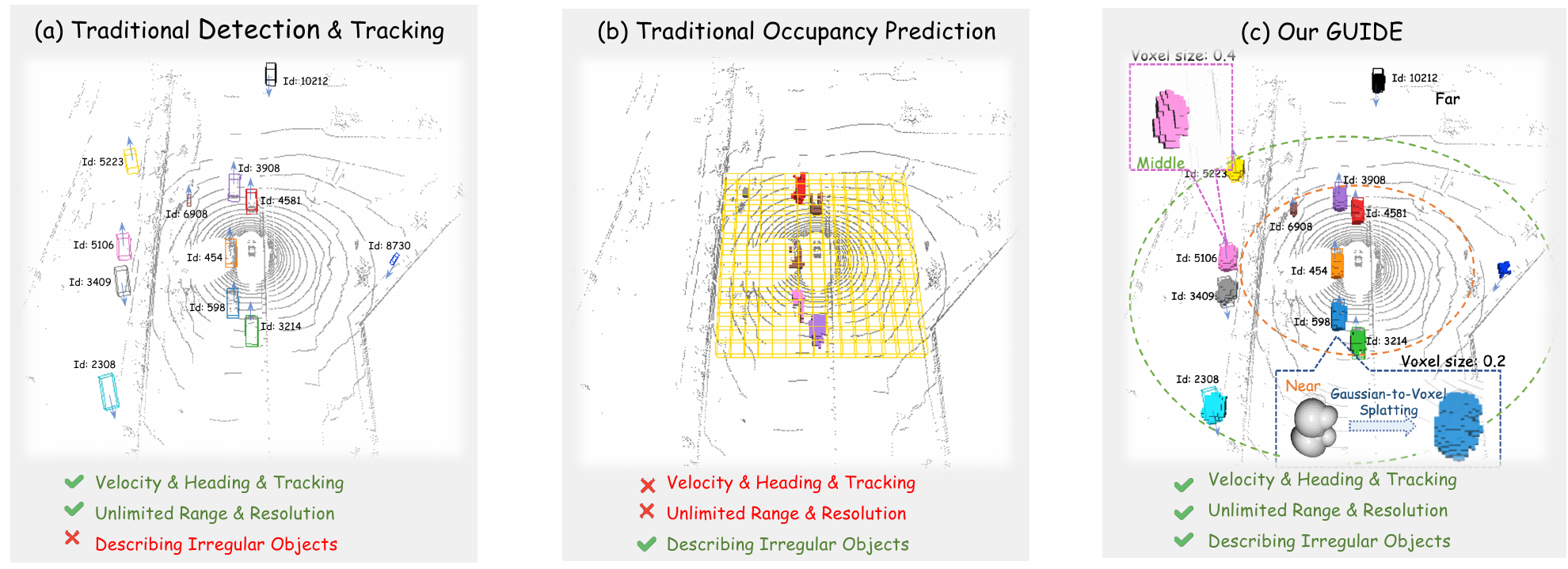}
  \captionof{figure}{\textbf{Comparison of traditional detection and tracking, traditional occupancy prediction and our proposed GUIDE.} The background points are only shown for scene comprehension and are not used in model inference. (a) Traditional detection and tracking methods output 3D bounding boxes equipped with directional and velocity information, and each box is associated with a unique ID for tracking purposes. (b) Traditional occupancy prediction interacts with voxel features, producing rough occupancy predictions, yet it lacks directional and velocity information. (c) GUIDE represents each instance using multiple 3D Gaussians enriched with directional and velocity information as well as tracking IDs, and generates high-quality instance occupancy predictions via Gaussian-to-Voxel Splatting.}
  \label{fig1}
\end{figure*}
}

\begin{abstract}

In the realm of autonomous driving, accurately detecting surrounding obstacles is crucial for effective decision-making. Traditional methods primarily rely on 3D bounding boxes to represent these obstacles, which often fail to capture the complexity of irregularly shaped, real-world objects. To overcome these limitations, we present GUIDE, a novel framework that utilizes 3D Gaussians for instance detection and occupancy prediction. Unlike conventional occupancy prediction methods, GUIDE also offers robust tracking capabilities. Our framework employs a sparse representation strategy, using Gaussian-to-Voxel Splatting to provide fine-grained, instance-level occupancy data without the computational demands associated with dense voxel grids. Experimental validation on the nuScenes dataset demonstrates GUIDE's performance, with an instance occupancy mAP of 21.61, marking a 50\% improvement over existing methods, alongside competitive tracking capabilities. GUIDE establishes a new benchmark in autonomous perception systems, effectively combining precision with computational efficiency to better address the complexities of real-world driving environments. The code is available at \url{https://github.com/CN-ADLab/GUIDE}.

\end{abstract}



\section{Introduction}

\label{sec:intro}

Accurate detection of surrounding obstacles is fundamental to autonomous driving systems. Despite rapid advancements in end-to-end driving technologies~\cite{hu2023planning, jiang2023vad, tong2023scene}, obstacle instance detection continues to be a critical component, also serving as an auxiliary task to expedite model convergence. Traditional methods for obstacle detection~\cite{huang2021bevdet, liang2022bevfusion, li2024bevformer} typically employ 3D bounding boxes to denote the position and size of objects. While effective for standard obstacles, this simplistic representation falls short in complex environments with diverse distributions. Real-world driving scenarios, featuring irregular obstacles like barriers, debris, billboards, and complex situations such as pedestrians carrying objects or cars with open doors, challenge the effectiveness of 3D bounding boxes.


To address these issues, occupancy prediction has emerged as a promising solution, providing a more flexible approach for describing complex-shaped obstacles by predicting the occupancy and category of each voxel grid~\cite{cao2022monoscene, huang2023tri, tian2023occ3d}. However, traditional occupancy methods~\cite{li2023voxformer, wei2023surroundocc, li2023fb, yu2023flashocc} rely on dense voxel features, resulting in memory usage that increases cubically with higher resolutions and broader perception ranges. Although some approaches~\cite{liu2024fully} attempt to mitigate these limitations by using coarse-to-fine strategies, the substantial memory overhead continues to restrict the practical application of these methods. Furthermore, most traditional occupancy approaches focus solely on the semantic category of voxel occupancy. While some methods, like SparseOcc~\cite{liu2024fully}, venture into predicting instance-level occupancy masks, they are confined to per-frame predictions, lacking the capability to estimate instance velocity or perform temporal tracking—vital components for autonomous driving decision-making. Consequently, real-world systems often require combining the outputs of 3D object detection and tracking tasks with occupancy results, relying heavily on manual post-processing to achieve final perception outputs.

In light of these challenges, we are motivated to design a unified framework that not only offers efficient instance-level occupancy prediction but also integrates detection and tracking. Such a holistic framework provides a more comprehensive and robust perception solution for autonomous driving systems. Recent advancements in 3D Gaussian representations offer a promising pathway for this unified approach. 3D Gaussians~\cite{kerbl20233d} have gained traction in scene reconstruction due to their flexible and sparse representation abilities. Building on this foundation, the GaussianFormer series~\cite{huang2024gaussianformer, huang2024probabilistic} employ sparse Gaussians, generating occupancy predictions through Gaussian-to-Voxel Splatting, thereby reducing dependency on dense voxel features and lowering computational demands. Despite these advances, GaussianFormer largely supports semantic occupancy predictions without instance-level specificity.

Inspired by these developments, we propose GUIDE, a novel Gaussian-based Unified Instance Detection framework. GUIDE uses multiple 3D Gaussians to represent each instance, employing Gaussian-to-Voxel Splatting for generating instance-level occupancy predictions. In addition, by leveraging Gaussian features, GUIDE outputs 3D bounding boxes and tracking IDs for each instance. This inherently sparse framework eliminates the necessity for dense voxel features, offering considerable memory efficiency. The continuous nature of Gaussian representations endows GUIDE with remarkable adaptability in modeling occupancy at varying resolutions. Specifically, the voxel size for occupancy prediction can be adjustably configured during inference, eliminating the need for retraining and facilitating efficient adaptation to diverse application requirements. We also introduce a novel mAP computation method tailored for assessing instance occupancy predictions. Experimental evaluations on the nuScenes~\cite{caesar2020nuscenes} dataset underscore our approach's effectiveness: GUIDE achieves an instance occupancy detection mAP of 21.61, showing a 50\% improvement over SparseOcc, while maintaining competitive performance in detection and tracking tasks.

In summary, our primary contributions are as follows:

\begin{itemize}
\item We propose GUIDE, a Gaussian-based unified instance detection framework that supports instance-level occupancy prediction while simultaneously performing traditional 3D object detection and tracking.

\item Our method employs a fully sparse representation for instance occupancy, greatly enhancing memory efficiency and allowing for flexible adjustment of inference resolution, thanks to the properties of Gaussian representation.

\item GUIDE achieves an instance occupancy detection mAP of 21.61 on the nuScenes benchmark, reflecting a 50\% improvement over SparseOcc, and delivers competitive performance in both detection and tracking tasks.
\end{itemize}
\begin{figure*}[t]
  \centering
   \includegraphics[width=1.0\linewidth]{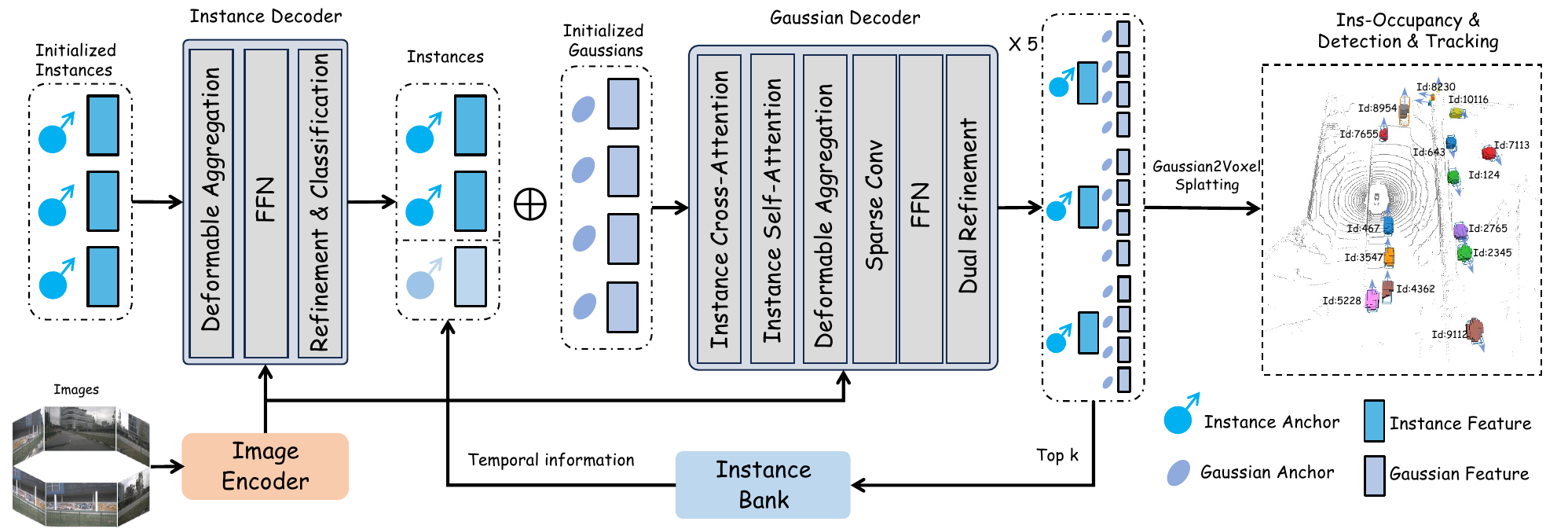}
   \caption{\textbf{Framework of our GUIDE.} 
    Instance queries and their anchors are subsequently initialized and iteratively updated through interactions with image features using the instance decoder.
   The updated top-k instances are combined with those in the historical instance bank to form a new candidate instance set. Each instance is then associated with multiple 3D Gaussians, which serve as their representations. These Gaussians are refined iteratively through a 5-layer Gaussian Decoder. Subsequently, instance occupancy predictions are generated via Gaussian-to-Voxel Splatting.
   And aggregating Gaussian features allows reconstruction of instance-level representations to predict each instance's bounding box and category. Additionally, the top-k instances update the instance bank, adding temporal information to aid inference for later frames. Meanwhile, we assign unique IDs to instances whose confidence scores exceed a predefined threshold in the instance bank for instance tracking across frames.
   }
   \label{fig2}
\end{figure*}
\section{Related Work}
\label{sec:related_work}

\subsection{Multi-view 3D Object Detection and Tracking}

The task of vision-based multi-view 3D object detection is crucial for autonomous driving systems that rely on visual sensors~\cite{park2021pseudo, wang2021fcos3d, wang2022detr3d}. Current methods in this field are typically categorized by whether they require dense Bird's Eye View (BEV) features. Among the methods that necessitate dense BEV features~\cite{li2023bevdepth, han2024exploring}, BEVDet~\cite{huang2021bevdet} is noteworthy. This approach utilizes the Lift-Splat-Shoot (LSS)~\cite{philion2020lift} technique, which projects image features into BEV space using depth estimation. Another method involves generating BEV queries that are mapped into image space to sample and construct BEV features~\cite{jiang2023polarformer, yang2023bevformer, li2024bevformer}.
Conversely, methods that do not require dense BEV features follow distinct technical paths. For instance, the PETR series~\cite{liu2022petr, liu2023petrv2, wang2023exploring} employs implicit 3D position encoding within object queries and image features, enabling direct prediction of 3D bounding boxes through global attention mechanisms. The Sparse4D series~\cite{lin2022sparse4d, lin2023sparse4d, lin2023sparse4dv3} utilizes explicit anchors to project and sample local features effectively for 3D object detection.
There has been a notable evolution from early approaches using CNN-based detection decoders to those utilizing query-based transformer decoders. Similarly, the instance tracking task has progressed from relying on dense features~\cite{hu2022monocular} to employing sparse queries~\cite{zhang2022mutr3d, gu2023vip3d} for tracking purposes. The inherent alignment between queries and instances in transformer decoders makes them well-suited for modular, end-to-end autonomous driving frameworks like UniAD~\cite{hu2023planning} and SparseDrive~\cite{sun2024sparsedrive}.

\subsection{Vision-based 3D Occupancy Prediction}

Recent advances in vision-based 3D occupancy prediction have enhanced autonomous driving perception, particularly for long-tail corner cases involving diverse or atypical obstacles~\cite{huang2023tri, tian2023occ3d}. MonoScene~\cite{cao2022monoscene} pioneered end-to-end monocular 3D semantic grid prediction using a 2D-3D UNet~\cite{ronneberger2015u}. With the rise of Bird's Eye View (BEV) representations, subsequent methods leverage BEV features to reconstruct dense voxel occupancies~\cite{yu2023flashocc, li2023fb, zhang2023occformer}, though often at high computational cost. To improve efficiency, coarse-to-fine upsampling~\cite{wang2024panoocc} and 2D rendering~\cite{pan2024renderocc} have been proposed, yet they suffer from information loss and rely heavily on accurate depth estimation. While most work focuses on semantic occupancy, recent efforts explore panoramic occupancy prediction, expanding the task’s scope~\cite{liu2024fully, moon2025mitigating}.


\subsection{Perception with 3D Gaussians}

The use of 3D Gaussian splatting has gained attention in scene reconstruction for its exceptional rendering quality and speed~\cite{kerbl20233d}. Recent studies highlight its potential in autonomous driving perception~\cite{lu2025toward}. GaussianBEV~\cite{chabot2025gaussianbev} achieves notable success in BEV segmentation by generating 3D Gaussians from image features and splatting them into BEV features. GaussianFormer~\cite{huang2024gaussianformer} pioneered Gaussian-to-Voxel Splatting, using 3D Gaussians for 3D semantic occupancy prediction, significantly improving efficiency. GaussianFormer-v2~\cite{huang2025gaussianformer} refines occupancy and semantic estimation within splatting, reducing Gaussian count. While the series offers memory efficiency over traditional methods via sparse representation, it only supports semantic — not instance-level — prediction. GaussianAD~\cite{zheng2024gaussianad} demonstrates 3D Gaussians' versatility across tasks like real-time mapping, further showcasing their potential in autonomous driving. 
\section{Proposed Approach}



In this work, we introduce GUIDE, a novel unified framework designed to enhance obstacle perception in autonomous driving systems. GUIDE processes surround-view images, employing 3D Gaussian representations to efficiently produce instance-level occupancy predictions for nearby obstacles, while also offering strong detection and tracking capabilities.

\subsection{Overall Architecture}

As illustrated in \cref{fig2}, GUIDE consists of four essential components: an image encoder, an instance decoder, a Gaussian decoder, and an instance bank.


The image encoder processes multi-view images represented as $I \in \mathbb{R}^{N \times H \times W \times 3}$, where $N$ is the number of surround-view cameras, and $H$ and $W$ are the image height and width, respectively. Leveraging a backbone and neck network, the image encoder extracts multi-scale features, denoted by $\mathbf{F} = \{ \mathbf{F}_s \}_{s=1}^{4}$, where $\mathbf{F}_s \in \mathbb{R}^{N \times H_s \times W_s \times C}$, with $s$ indicating the feature scales and $C$ denoting the number of feature channels.

The instance decoder utilizes the extracted features and initialized anchors to identify individual object instances, while the Gaussian decoder employs multiple 3D Gaussian representations for each instance to accurately model spatial occupancy. Aggregating Gaussian features enables the reconstruction of instance-level representations for category and 3D bounding box prediction. To address the adverse impact of partial object observation in single-frame images, we introduce an instance bank to enhance temporal feature fusion across sequences. Specifically, historical instance features are first concatenated with the outputs of the instance decoder, yielding a more comprehensive set of candidate instances. Next, in the Gaussian decoder, these historical features dynamically interact with current instance candidates, providing essential contextual information that improves the precision of occupancy predictions. Additionally, the instance bank assigns unique IDs to instances with confidence scores exceeding a predefined threshold~\cite{lin2023sparse4dv3}, facilitating instance tracking across frames.



\subsection{Instance Decoder}

In this module, instances are represented by a set of features \(\mathbf{F}_I \in \mathbb{R}^{N_I \times C}\) and anchors \(\mathbf{B}_I \in \mathbb{R}^{N_I \times 10}\). \(N_I\) denotes the number of candidate instances, and \(C\) represents the number of channels for the instance features. Each anchor comprises a 10-dimensional vector encoding the position, scale, yaw angle, and velocity. Instance features are initialized to zeros, while anchors are initialized using k-means clustering over the ground truth data from the training set. Then the information of each instance's anchor is encoded into \(\mathbf{E}_I \in \mathbb{R}^{N_I \times C}\), which serves as the positional encoding for the instance features during attention operations.

The instance decoder consists of three modules: deformable aggregation, a feedforward network (FFN), and instance refinement. With deformable aggregation, each instance anchor center is projected onto the multi-view image features, generating both fixed and learnable offsets for feature sampling around the projected center. Image features aggregated from these sampling points are used to update the corresponding instance features. The updated features are further processed by the FFN and then passed to the instance refinement module, which predicts instance classes and anchor offset residuals to refine the anchors.

\begin{table*}[tbp]
\centering
\begin{tabular}{@{}c|c|c|c|c|c|c|c|>{\columncolor{lightgray}}cc@{}}
\hline
{\textbf{Methods}} & {\textbf{Backbone}} & {\textbf{Input Size}} & {\textbf{mAP↑}} & {\textbf{NDS↑}} & {\textbf{AMOTA↑}} & 
{\textbf{IDS$\downarrow$}} & {\textbf{mIOU$^{10}$↑}} & {\textbf{mAP$_{occ}^{8}$↑}} & {\textbf{mAP$_{occ}^{10}$↑}} 
\\
\hline
DETR3D     & R101 & 1333x800 & 30.3 & 37.4 & - & - & - & - & - \\
PETR     & R50 & 1056x384 & 31.3 & 38.1 & - & - & - & - & -\\
BevFormer & R101 & 1600x900 & \textbf{44.5} & \textbf{53.5} & - & - & - & - & -\\
\hline
ViP3D & R50 & 1333x800 & - & - & 21.7 & - & - & - & -\\
QD-3DT & R101 & 1600x900 & - & - & 24.2 & - & - & - & -\\
MUTR3D & R101 & 1600x900 & - & - & 29.4 & 3822 & - & - & -\\
\hline
UniAD & R101 & 1600x900 & 38.0 & 49.8 & 35.9 & 906 & - & - & -\\
SparseDrive-S & R50 & 704x256 & 41.8 & 52.5 & 38.6 & 886 & - & - & -\\
\hline
RenderOcc     & Swin-B & 1408×512 & - & - & - & - & 17.30 & - & - \\
OccFormer     & R50 & 704x256 & - & - & - & - & 21.38 & - & - \\
CTF-Occ & R101 & 960x640 & - & - & - & - & 27.97 & - & - \\
GaussianFormer   & R101 & 1600x864 & - & - & - & - & \textbf{29.93} & - & - \\
\hline
SparseOCC & R50 & 704x256 & - & - & - & - & 25.31 & 14.40 & -   \\
\textbf{GUIDE}     & R50 & 704x256 & 41.0 & 51.8 & \textbf{39.4} & \textbf{516} & 25.39 & \textbf{21.61} & 22.81  \\
\hline
\end{tabular}
\caption{\textbf{Performance comparison on the nuScenes val set.} The metric mIOU$^{10}$ represents the mean Intersection over Union for all 10 foreground categories in the 3D semantic occupancy evaluation. The mAP$_{occ}^{8}$ metric indicates the average instance occupancy mean Average Precision across thresholds $\{0.1, 0.2, 0.3\}$ for 8 categories, excluding the general obstacle category. Meanwhile, mAP$_{occ}^{10}$ reflects the same evaluation performed across all 10 foreground categories.}\label{tbl1}
\end{table*}

\subsection{Gaussian Decoder}

The top-k instances from the instance decoder, along with their corresponding features and anchors, are combined with those from the historical instance bank to form a new candidate instance set. In the Gaussian decoder, each instance is associated with a set of 3D Gaussians, parameterized by learnable Gaussian features \(\mathbf{F}_{gs} \in \mathbb{R}^{K \times C}\) and Gaussian anchors \(\mathbf{G} \in \mathbb{R}^{K \times 10}\), where \(K\) denotes the number of Gaussians per instance, and \(C\) s the feature dimension. Each Gaussian anchor encodes the offset from the instance center to the Gaussian mean, the scale, and the rotation quaternion. These initialized Gaussians are combined with candidate instances to produce instance-level Gaussian features \(\mathbf{F}_{Igs} \in \mathbb{R}^{N_I \times K \times C}\) and Gaussian anchors \(\mathbf{G}_I \in \mathbb{R}^{N_I \times K \times 10}\). The instance features could be obtained by averaging the Gaussian features for each instance.

The Gaussian decoder contains five layers. To enhance contextual information, each layer first performs cross-attention between instance features and those in the historical instance bank, followed by self-attention among the instance features. The updated instance features are then used to refine the corresponding Gaussian features. Each Gaussian feature, treated as a separate query, is further updated by the deformable aggregation module, which projects Gaussian anchors onto the image feature maps and samples relevant features. These aggregated features are combined with the Gaussian features, which are then processed by sparse 3D convolutions for local self-encoding.

In the dual refinement module, Gaussian and instance anchors are jointly updated using both Gaussian and instance features, where the instance features are computed by averaging their corresponding Gaussian features. The refined features and anchors are passed to the next layer of the Gaussian decoder. In the final layer, the features and anchors of the top-k instances are used to update the historical instance bank.

\subsection{Instance Gaussian Splatting}

Inspired by the GaussianFormer series~\cite{huang2024gaussianformer, huang2025gaussianformer}, our approach employs a Gaussian-to-Voxel Splatting technique to predict 3D instance occupancy by splatting instance-level 3D Gaussians. Unlike GaussianFormer~\cite{huang2024gaussianformer}, our method only performs geometric occupancy splatting, omitting semantic splatting. Instead, semantic labels for instance occupancy are directly obtained from the instance classification results. In this process, each 3D Gaussian represents a probabilistic spatial occupancy, where the probability at the center \(m\) is set to 1 and decreases with distance according to the 3D Gaussian distribution.

\begin{equation}
  p(x; G) = \exp\left(-\frac{1}{2} (x-m)^T \Sigma^{-1} (x-m)\right)
\end{equation}
\begin{equation}
\Sigma = R S S^T R^T
\end{equation}

As illustrated in the equations above, \(p(x; G)\) denotes the probability that point \(x\) is occupied as determined by 3D Gaussian \(G\). Here, \(\Sigma\) is the covariance matrix, \(R\) the rotation matrix, and \(S\) the diagonal scale matrix.


In the process of splatting all Gaussians for each instance, the probability that a point is occupied by different Gaussians is considered to be independent. By applying the multiplication rule of probabilities, the aggregated probability that a point in space is occupied can be computed as follows.

\begin{equation}
p(x) = 1 - \prod_{i=1}^{K} \left(1 - p(x; G_i)\right)
\end{equation}

\subsection{Training Loss}

During training, the supervised loss is composed of three components: regression, classification and occupancy. 

\begin{equation}
\mathcal{L} = \mathcal{L}_{\text{reg}} + \mathcal{L}_{\text{cls}} + \mathcal{L}_{\text{occ}}
\end{equation}

We adopt the Hungarian matching algorithm to associate predicted instances with ground truth instances. Specifically, the regression tasks, which predict instance center positions, scales, and velocities, are supervised using the L1 loss function. For the instance classification tasks, we incorporate a multi-class focal loss. For occupancy, we also employ focal loss~\cite{lin2017focal} to address the class imbalance between occupied and unoccupied voxels in the ground truth.

\begin{figure*}[t]
    \centering
    \includegraphics[width=1.0\textwidth]{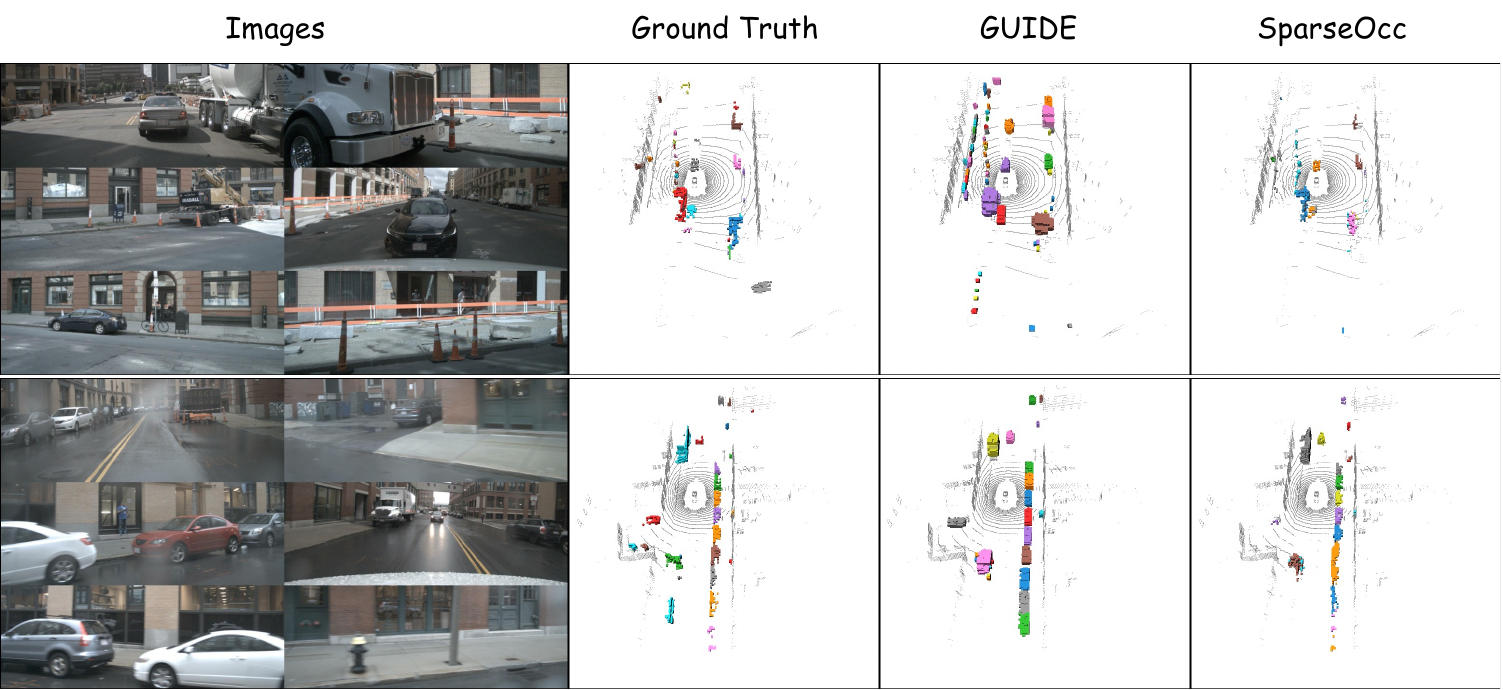}
    \caption{\textbf{Visualization results for 3D instance occupancy prediction on nuScenes.} The different colors are used to represent the occupancy predictions for various instances. The background points are included to aid scene comprehension and are not utilized during the model inference process.}
   \label{fig3}
\end{figure*}

\section{Experiments}

\subsection{Datasets}
Our experiments are conducted on the nuScenes~\cite{caesar2020nuscenes} dataset, which contains 1000 sequences officially split into 750 for training, 150 for validation, and 150 for testing. Each sequence spans 20 seconds and provides keyframe detections and tracking ground truth at 2 Hz, along with six surround-view images.

To obtain occupancy ground truth, we use the annotations provided by Occ3D~\cite{tian2023occ3d}, which supplies occupancy labels for the nuScenes dataset in the ego vehicle coordinate system. The annotations cover a range of $[-40\,\mathrm{m}, -40\,\mathrm{m}, -1\,\mathrm{m},\ 40\,\mathrm{m},\ 40\,\mathrm{m},\ 5.4\,\mathrm{m}]$ with a voxel resolution of $[0.4\,\mathrm{m},\ 0.4\,\mathrm{m},\ 0.4\,\mathrm{m}]$. The semantic labels in Occ3D are consistent with those used in nuScenes LiDAR segmentation tasks. To support foreground instance detection in our study, we further utilize tools from SparseOcc~\cite{liu2024fully} to generate instance-level occupancy ground truth for foreground objects with IDs.


\subsection{Evaluation Metrics}

To our knowledge, GUIDE is the first method to address foreground instance occupancy prediction. For evaluation, we adopt a metric inspired by the mean Average Precision (mAP) commonly used in object detection. Specifically, we use the Intersection over Union (IoU) between predicted and ground-truth instance occupancies as the thresholding criterion, enabling finer-grained measurement. To comprehensively assess both the precision and recall of instance occupancy predictions, we compute mAP at multiple IoU thresholds \{0.1, 0.2, 0.3\}, and report the average as our final evaluation metric.
\begin{equation}
    mAP_{occ} = Average(mAP_{0.1}, mAP_{0.2}, mAP_{0.3})
\end{equation}

Additionally, we employ the commonly used mean Intersection over Union (mIoU) metric from the 3D semantic occupancy task as a supplementary evaluation measure, which allows us to indirectly assess the semantic occupancy prediction capability of our method. Furthermore, we utilize the standard mAP and NDS metrics on the nuScenes dataset for detection performance evaluation, as well as AMOTA and IDS for tracking evaluation.



\subsection{Main Results}
In this section, we evaluate the performance of the proposed GUIDE on the nuScenes validation set. To the best of our knowledge, GUIDE is the first framework to address instance occupancy prediction focusing exclusively on the foreground categories in nuScenes. The most closely related prior work is SparseOcc, which employs a mask transformer for panoramic occupancy prediction on this dataset. However, there are key distinctions between the two approaches. SparseOcc performs instance-level predictions only for eight foreground categories, such as cars and pedestrians, and provides semantic occupancy predictions for the remaining classes. In contrast, our method extends instance occupancy prediction to additional obstacle categories, specifically barriers and traffic cones in nuScenes. This extension enables unified instance detection for all foreground obstacles.

\begin{table}[tbp]
\setlength{\tabcolsep}{5pt} 
\centering
\begin{tabular}{@{}c|cc|c|c@{}}
\hline
\textbf{Methods} & \textbf{barrier} & \textbf{traffic cone} & \textbf{mIOU↑} & \textbf{mAP$_{occ}$↑} \\
\hline
SparseOCC & 33.59 & \textbf{24.39} & 28.99 & - \\
\textbf{GUIDE}     & \textbf{39.14} & 22.78 & \textbf{30.96} & \textbf{27.79} \\
\hline
\end{tabular}
\caption{\textbf{Performance for general obstacles on the nuScenes val set.} The mIOU metric represents the average mean IOU for the barrier and traffic cone categories, based on the 3D semantic occupancy evaluation. Similarly, the mAP metric is evaluated exclusively for them.}\label{tbl2}
\end{table}
\begin{table}[tbp]
\setlength{\tabcolsep}{5pt} 
\centering
\begin{tabular}{@{} c|*{3}{w{c}{1.75cm}} @{}}
\hline
{$\textbf{Level}_{crowd}$↑} & \textbf{Low}($\leq 20$) & \textbf{Medium}(20-40) & \textbf{High}($\geq 40$) \\
\hline
GUIDE  & 32.35 & 28.16 & 31.92 \\
\hline
\end{tabular}
\caption{\textbf{Performance of mAP$_{occ}$ at IoU threshold 0.1 under different Crowding Levels.} The level is linked to the instance number per scene.}\label{tbl12}
\end{table}


To evaluate the performance of the proposed GUIDE on instance occupancy prediction, we conduct a comparative analysis with SparseOcc using the proposed instance occupancy mAP metric. This evaluation focuses on the eight foreground categories, including cars and pedestrians. As shown in \cref{tbl1}, GUIDE achieves an average instance occupancy mAP of 21.61 for these categories, representing a 50\% improvement over SparseOcc. For all foreground categories, including general obstacles, GUIDE attains a mean instance occupancy mAP of 22.81. These results indicate that GUIDE demonstrates exceptional accuracy and recall in instance occupancy prediction. We attribute this performance to the instance-centric design of our method, which progressively refines spatial information at the instance level and thus achieves higher boundary precision in occupancy prediction.

\begin{figure*}
  \centering
   \includegraphics[width=1.0\linewidth]{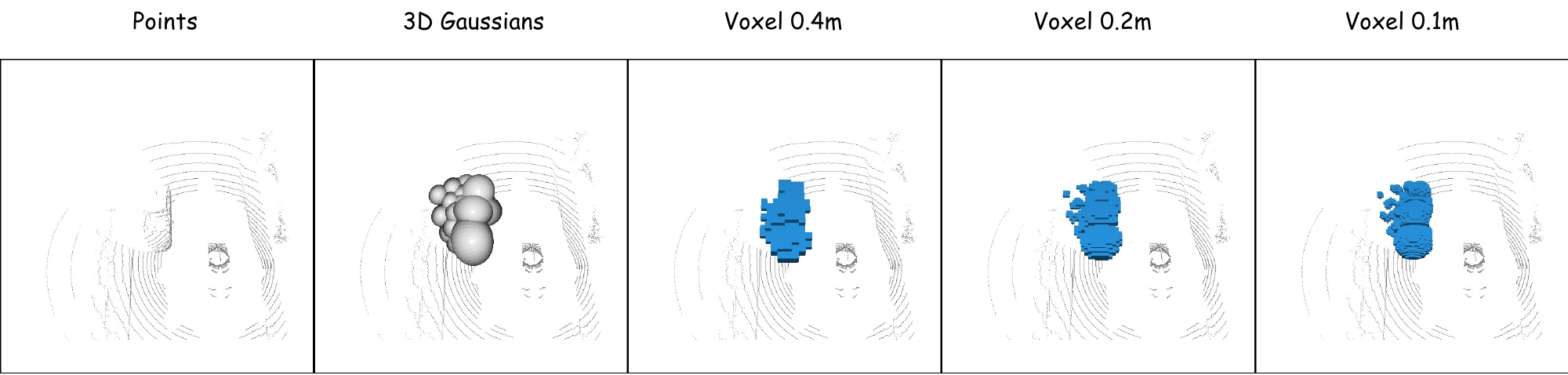}
   \caption{\textbf{Visualization for occupancy with different voxel sizes when processing the Gaussian-to-Voxel Splatting.} }
   \label{fig4}
\end{figure*}



\begin{table}[tbp]
\centering
\begin{tabular}[b]{c|c|cc}
\hline
\textbf{Methods}  & \textbf{Memory}  & \textbf{mIOU$^{10}$}↑ & \textbf{mAP$_{occ}^{8}$}↑ \\
\hline
SparseOcc & {$5424$ MB} & {$25.31$} & {$14.40$} \\
\textbf{GUIDE} & \textbf{3434 MB}  & \textbf{25.39}   & \textbf{21.61} \\
\hline
\end{tabular}
\caption{\textbf{Comparison with SparseOcc in terms of computation and memory.}}\label{tbl5}
\end{table}


Furthermore, to indirectly assess the semantic occupancy prediction capability of our method, we aggregate instance occupancy predictions according to their semantic categories. Since our method focuses on foreground categories, we recalculate the mIoU of the baselines for foreground classes alone to ensure a fair comparison. As shown in \cref{tbl1}, our method achieves competitive performance even with a smaller backbone and lower-resolution inputs. Moreover, with a conservative allocation of only 48 Gaussians per instance, GUIDE outperforms our primary baseline, SparseOcc, in semantic occupancy accuracy, despite both being instance occupancy approaches.

Distinctively, GUIDE overcomes the limitations of prior methods that are restricted solely to occupancy prediction, as the intrinsic properties of 3D Gaussian representations naturally endow our framework with integrated detection and tracking capabilities. As shown in \cref{tbl1}, GUIDE achieves competitive detection performance even with a relatively small network scale and low input resolution. Notably, GUIDE also exhibits improved multi-object tracking performance, particularly in terms of IDS, which primarily assesses the consistency of object tracking across frames. This notable gain in tracking stability may be attributed to the richer geometric and spatial information inherently encoded in the Gaussian representation, which facilitates a more stable and robust matching process during tracking.

In addition, \cref{tbl2} showcases the performance of GUIDE in predicting the semantic and instance-level occupancy of general obstacles. For semantic occupancy prediction, GUIDE achieves an mIoU of 30.96 for barriers and traffic cones, representing an improvement of 6.8\% over SparseOcc. This result underscores GUIDE's ability to accurately segment these common obstacle categories. Furthermore, for instance-level occupancy prediction of general obstacles, GUIDE achieves an mAP of 27.79, highlighting its capability to precisely identify and predict the occupancy of such instances.

We further evaluate the instance discrimination ability of GUIDE under varying crowding levels. As shown in \cref{tbl12}, performance remains stable even in dense scenarios, confirming its robustness to high instance density.

\subsection{Visualizations}


As illustrated in \cref{fig3}, we present the instance occupancy prediction results of GUIDE across various scenes, alongside a comparison with SparseOcc. The figure shows that, compared to SparseOcc, GUIDE produces significantly denser instance occupancy predictions. Moreover, GUIDE demonstrates superior instance recall capability. Even in scenarios with numerous targets, GUIDE maintains high accuracy while achieving outstanding recall performance.



Moreover, we visualize Gaussian-to-Voxel results at different resolutions. Leveraging the continuous nature of the Gaussian representation, GUIDE enables flexible adjustment of the predicted instance occupancy resolution at inference time without retraining. As illustrated in \cref{fig4}, reducing the voxel size leads to enhanced detail and precision in the occupancy prediction results. This demonstrates that our method offers strong flexibility in resolution. In practical autonomous driving systems, the splatting resolution for instance occupancy during inference can be dynamically adjusted as needed. For example, high resolution such as 0.1\,m can be used for nearby instances requiring precise localization, while lower resolutions such as 0.4\,m can be applied to distant targets.



\begin{table}[tbp]
\centering
\begin{tabular}{>{\centering\arraybackslash}c|>
{\centering\arraybackslash}c|>{\centering\arraybackslash}c>{\centering\arraybackslash}c}
\hline
\textbf{Gaussian Number} & \textbf{Latency$\downarrow$} & \textbf{mIOU↑} & \textbf{mAP$_{occ}$↑} \\
\hline
16 & \textbf{205ms} & 22.79 & 18.94 \\
32 & 291ms & 23.31 & 20.21\\
48 & 373ms & \textbf{25.39} & \textbf{22.81} \\
\hline
\end{tabular}
\caption{\textbf{Ablation on the number of Gaussians for each instance.} Both the mIOU and mAP metrics reflect evaluations conducted across all 10 foreground categories.}\label{tbl3}
\end{table}

\subsection{Memory Efficiency}

We further compare the memory efficiency of GUIDE with SparseOcc. The experiments are conducted on a single NVIDIA 3090 GPU with a batch size of 1. As shown in \cref{tbl5}, GUIDE achieves a 36.7\% reduction in GPU memory consumption during inference compared to SparseOcc. This result demonstrates the advantage of our sparse Gaussian representation in terms of memory efficiency.



\subsection{Ablation Study}

In this section, we investigate the effect of varying the number of Gaussians used to model each instance on the performance of GUIDE. As reported in \cref{tbl3}, our empirical results show that increasing the number of Gaussians per instance consistently enhances the accuracy of instance occupancy predictions. This improvement can be attributed to the model's greater capacity to represent complex spatial distributions with higher precision.

\section{Conclusion}
In this work, we propose GUIDE, a novel Gaussian-based unified instance detection framework that employs 3D Gaussians as flexible and efficient intermediate representations for autonomous vehicle perception. By replacing dense voxel grids with Gaussian-to-Voxel Splatting, GUIDE substantially reduces computational and memory demands while enhancing the accuracy of instance occupancy prediction. Furthermore, GUIDE also supports both detection and tracking tasks, and achieves competitive performance. Comprehensive experiments on the nuScenes benchmark demonstrate that GUIDE outperforms the previous instance occupancy method, SparseOcc, in both memory efficiency and occupancy detection accuracy. These findings illustrate the promise of GUIDE as a robust and scalable solution for perception in end-to-end autonomous driving systems.

\section*{Acknowledgments}

This research was supported by the Zhejiang Provincial Natural Science Foundation of China under Grant No. LD24F030001.

\bibliography{main}

@String(AAAI = {AAAI})

@inproceedings{zhang2023occformer,
  title={Occformer: Dual-path transformer for vision-based 3d semantic occupancy prediction},
  author={Zhang, Yunpeng and Zhu, Zheng and Du, Dalong},
  booktitle={Proceedings of the IEEE/CVF International Conference on Computer Vision},
  pages={9433--9443},
  year={2023}
}

@inproceedings{pan2024renderocc,
  title={Renderocc: Vision-centric 3d occupancy prediction with 2d rendering supervision},
  author={Pan, Mingjie and Liu, Jiaming and Zhang, Renrui and Huang, Peixiang and Li, Xiaoqi and Xie, Hongwei and Wang, Bing and Liu, Li and Zhang, Shanghang},
  booktitle={2024 IEEE International Conference on Robotics and Automation (ICRA)},
  pages={12404--12411},
  year={2024},
  organization={IEEE}
}

@inproceedings{lin2017feature,
  title={Feature pyramid networks for object detection},
  author={Lin, Tsung-Yi and Doll{\'a}r, Piotr and Girshick, Ross and He, Kaiming and Hariharan, Bharath and Belongie, Serge},
  booktitle={Proceedings of the IEEE conference on computer vision and pattern recognition},
  pages={2117--2125},
  year={2017}
}

@inproceedings{he2016deep,
  title={Deep residual learning for image recognition},
  author={He, Kaiming and Zhang, Xiangyu and Ren, Shaoqing and Sun, Jian},
  booktitle={Proceedings of the IEEE conference on computer vision and pattern recognition},
  pages={770--778},
  year={2016}
}

@article{zheng2024gaussianad,
  title={GaussianAD: Gaussian-Centric End-to-End Autonomous Driving},
  author={Zheng, Wenzhao and Wu, Junjie and Zheng, Yao and Zuo, Sicheng and Xie, Zixun and Yang, Longchao and Pan, Yong and Hao, Zhihui and Jia, Peng and Lang, Xianpeng and others},
  journal={arXiv preprint arXiv:2412.10371},
  year={2024}
}

@inproceedings{wang2024panoocc,
  title={Panoocc: Unified occupancy representation for camera-based 3d panoptic segmentation},
  author={Wang, Yuqi and Chen, Yuntao and Liao, Xingyu and Fan, Lue and Zhang, Zhaoxiang},
  booktitle={Proceedings of the IEEE/CVF conference on computer vision and pattern recognition},
  pages={17158--17168},
  year={2024}
}

@inproceedings{ronneberger2015u,
  title={U-net: Convolutional networks for biomedical image segmentation},
  author={Ronneberger, Olaf and Fischer, Philipp and Brox, Thomas},
  booktitle={Medical image computing and computer-assisted intervention--MICCAI 2015: 18th international conference, Munich, Germany, October 5-9, 2015, proceedings, part III 18},
  pages={234--241},
  year={2015},
  organization={Springer}
}

@article{sun2024sparsedrive,
  title={Sparsedrive: End-to-end autonomous driving via sparse scene representation},
  author={Sun, Wenchao and Lin, Xuewu and Shi, Yining and Zhang, Chuang and Wu, Haoran and Zheng, Sifa},
  journal={arXiv preprint arXiv:2405.19620},
  year={2024}
}

@article{lin2023sparse4dv3,
  title={Sparse4d v3: Advancing end-to-end 3d detection and tracking},
  author={Lin, Xuewu and Pei, Zixiang and Lin, Tianwei and Huang, Lichao and Su, Zhizhong},
  journal={arXiv preprint arXiv:2311.11722},
  year={2023}
}

@article{lin2023sparse4d,
  title={Sparse4d v2: Recurrent temporal fusion with sparse model},
  author={Lin, Xuewu and Lin, Tianwei and Pei, Zixiang and Huang, Lichao and Su, Zhizhong},
  journal={arXiv preprint arXiv:2305.14018},
  year={2023}
}

@article{lin2022sparse4d,
  title={Sparse4d: Multi-view 3d object detection with sparse spatial-temporal fusion},
  author={Lin, Xuewu and Lin, Tianwei and Pei, Zixiang and Huang, Lichao and Su, Zhizhong},
  journal={arXiv preprint arXiv:2211.10581},
  year={2022}
}

@inproceedings{wang2023exploring,
  title={Exploring object-centric temporal modeling for efficient multi-view 3d object detection},
  author={Wang, Shihao and Liu, Yingfei and Wang, Tiancai and Li, Ying and Zhang, Xiangyu},
  booktitle={Proceedings of the IEEE/CVF international conference on computer vision},
  pages={3621--3631},
  year={2023}
}

@inproceedings{liu2023petrv2,
  title={Petrv2: A unified framework for 3d perception from multi-camera images},
  author={Liu, Yingfei and Yan, Junjie and Jia, Fan and Li, Shuailin and Gao, Aqi and Wang, Tiancai and Zhang, Xiangyu},
  booktitle={Proceedings of the IEEE/CVF International Conference on Computer Vision},
  pages={3262--3272},
  year={2023}
}

@inproceedings{liu2022petr,
  title={Petr: Position embedding transformation for multi-view 3d object detection},
  author={Liu, Yingfei and Wang, Tiancai and Zhang, Xiangyu and Sun, Jian},
  booktitle={European conference on computer vision},
  pages={531--548},
  year={2022},
  organization={Springer}
}

@inproceedings{yang2023bevformer,
  title={Bevformer v2: Adapting modern image backbones to bird's-eye-view recognition via perspective supervision},
  author={Yang, Chenyu and Chen, Yuntao and Tian, Hao and Tao, Chenxin and Zhu, Xizhou and Zhang, Zhaoxiang and Huang, Gao and Li, Hongyang and Qiao, Yu and Lu, Lewei and others},
  booktitle={Proceedings of the IEEE/CVF Conference on Computer Vision and Pattern Recognition},
  pages={17830--17839},
  year={2023}
}

@inproceedings{jiang2023polarformer,
  title={Polarformer: Multi-camera 3d object detection with polar transformer},
  author={Jiang, Yanqin and Zhang, Li and Miao, Zhenwei and Zhu, Xiatian and Gao, Jin and Hu, Weiming and Jiang, Yu-Gang},
  booktitle={Proceedings of the AAAI conference on Artificial Intelligence},
  volume={37},
  number={1},
  pages={1042--1050},
  year={2023}
}

@article{han2024exploring,
  title={Exploring recurrent long-term temporal fusion for multi-view 3d perception},
  author={Han, Chunrui and Yang, Jinrong and Sun, Jianjian and Ge, Zheng and Dong, Runpei and Zhou, Hongyu and Mao, Weixin and Peng, Yuang and Zhang, Xiangyu},
  journal={IEEE Robotics and Automation Letters},
  year={2024},
  publisher={IEEE}
}

@inproceedings{li2023bevdepth,
  title={Bevdepth: Acquisition of reliable depth for multi-view 3d object detection},
  author={Li, Yinhao and Ge, Zheng and Yu, Guanyi and Yang, Jinrong and Wang, Zengran and Shi, Yukang and Sun, Jianjian and Li, Zeming},
  booktitle={Proceedings of the AAAI conference on artificial intelligence},
  volume={37},
  number={2},
  pages={1477--1485},
  year={2023}
}

@inproceedings{philion2020lift,
  title={Lift, splat, shoot: Encoding images from arbitrary camera rigs by implicitly unprojecting to 3d},
  author={Philion, Jonah and Fidler, Sanja},
  booktitle={Computer Vision--ECCV 2020: 16th European Conference, Glasgow, UK, August 23--28, 2020, Proceedings, Part XIV 16},
  pages={194--210},
  year={2020},
  organization={Springer}
}

@inproceedings{wang2022detr3d,
  title={Detr3d: 3d object detection from multi-view images via 3d-to-2d queries},
  author={Wang, Yue and Guizilini, Vitor Campagnolo and Zhang, Tianyuan and Wang, Yilun and Zhao, Hang and Solomon, Justin},
  booktitle={Conference on Robot Learning},
  pages={180--191},
  year={2022},
  organization={PMLR}
}

@inproceedings{wang2021fcos3d,
  title={Fcos3d: Fully convolutional one-stage monocular 3d object detection},
  author={Wang, Tai and Zhu, Xinge and Pang, Jiangmiao and Lin, Dahua},
  booktitle={Proceedings of the IEEE/CVF International Conference on Computer Vision},
  pages={913--922},
  year={2021}
}

@inproceedings{park2021pseudo,
  title={Is pseudo-lidar needed for monocular 3d object detection?},
  author={Park, Dennis and Ambrus, Rares and Guizilini, Vitor and Li, Jie and Gaidon, Adrien},
  booktitle={Proceedings of the IEEE/CVF International Conference on Computer Vision},
  pages={3142--3152},
  year={2021}
}

@inproceedings{caesar2020nuscenes,
  title={nuscenes: A multimodal dataset for autonomous driving},
  author={Caesar, Holger and Bankiti, Varun and Lang, Alex H and Vora, Sourabh and Liong, Venice Erin and Xu, Qiang and Krishnan, Anush and Pan, Yu and Baldan, Giancarlo and Beijbom, Oscar},
  booktitle={Proceedings of the IEEE/CVF conference on computer vision and pattern recognition},
  pages={11621--11631},
  year={2020}
}

@article{huang2024probabilistic,
  title={Probabilistic Gaussian Superposition for Efficient 3D Occupancy Prediction},
  author={Huang, Yuanhui and Thammatadatrakoon, Amonnut and Zheng, Wenzhao and Zhang, Yunpeng and Du, Dalong and Lu, Jiwen},
  journal={arXiv preprint arXiv:2412.04384},
  year={2024}
}

@inproceedings{huang2024gaussianformer,
  title={Gaussianformer: Scene as gaussians for vision-based 3d semantic occupancy prediction},
  author={Huang, Yuanhui and Zheng, Wenzhao and Zhang, Yunpeng and Zhou, Jie and Lu, Jiwen},
  booktitle={European Conference on Computer Vision},
  pages={376--393},
  year={2024},
  organization={Springer}
}

@article{kerbl20233d,
  title={3d gaussian splatting for real-time radiance field rendering.},
  author={Kerbl, Bernhard and Kopanas, Georgios and Leimk{\"u}hler, Thomas and Drettakis, George},
  journal={ACM Trans. Graph.},
  volume={42},
  number={4},
  pages={139--1},
  year={2023}
}

@inproceedings{liu2024fully,
  title={Fully sparse 3d occupancy prediction},
  author={Liu, Haisong and Chen, Yang and Wang, Haiguang and Yang, Zetong and Li, Tianyu and Zeng, Jia and Chen, Li and Li, Hongyang and Wang, Limin},
  booktitle={European Conference on Computer Vision},
  pages={54--71},
  year={2024},
  organization={Springer}
}

@article{yu2023flashocc,
  title={Flashocc: Fast and memory-efficient occupancy prediction via channel-to-height plugin},
  author={Yu, Zichen and Shu, Changyong and Deng, Jiajun and Lu, Kangjie and Liu, Zongdai and Yu, Jiangyong and Yang, Dawei and Li, Hui and Chen, Yan},
  journal={arXiv preprint arXiv:2311.12058},
  year={2023}
}

@article{li2023fb,
  title={Fb-occ: 3d occupancy prediction based on forward-backward view transformation},
  author={Li, Zhiqi and Yu, Zhiding and Austin, David and Fang, Mingsheng and Lan, Shiyi and Kautz, Jan and Alvarez, Jose M},
  journal={arXiv preprint arXiv:2307.01492},
  year={2023}
}

@inproceedings{wei2023surroundocc,
  title={Surroundocc: Multi-camera 3d occupancy prediction for autonomous driving},
  author={Wei, Yi and Zhao, Linqing and Zheng, Wenzhao and Zhu, Zheng and Zhou, Jie and Lu, Jiwen},
  booktitle={Proceedings of the IEEE/CVF International Conference on Computer Vision},
  pages={21729--21740},
  year={2023}
}

@inproceedings{li2023voxformer,
  title={Voxformer: Sparse voxel transformer for camera-based 3d semantic scene completion},
  author={Li, Yiming and Yu, Zhiding and Choy, Christopher and Xiao, Chaowei and Alvarez, Jose M and Fidler, Sanja and Feng, Chen and Anandkumar, Anima},
  booktitle={Proceedings of the IEEE/CVF conference on computer vision and pattern recognition},
  pages={9087--9098},
  year={2023}
}

@article{tian2023occ3d,
  title={Occ3d: A large-scale 3d occupancy prediction benchmark for autonomous driving},
  author={Tian, Xiaoyu and Jiang, Tao and Yun, Longfei and Mao, Yucheng and Yang, Huitong and Wang, Yue and Wang, Yilun and Zhao, Hang},
  journal={Advances in Neural Information Processing Systems},
  volume={36},
  pages={64318--64330},
  year={2023}
}

@inproceedings{huang2023tri,
  title={Tri-perspective view for vision-based 3d semantic occupancy prediction},
  author={Huang, Yuanhui and Zheng, Wenzhao and Zhang, Yunpeng and Zhou, Jie and Lu, Jiwen},
  booktitle={Proceedings of the IEEE/CVF conference on computer vision and pattern recognition},
  pages={9223--9232},
  year={2023}
}

@inproceedings{cao2022monoscene,
  title={Monoscene: Monocular 3d semantic scene completion},
  author={Cao, Anh-Quan and De Charette, Raoul},
  booktitle={Proceedings of the IEEE/CVF Conference on Computer Vision and Pattern Recognition},
  pages={3991--4001},
  year={2022}
}

@article{huang2021bevdet,
  title={Bevdet: High-performance multi-camera 3d object detection in bird-eye-view},
  author={Huang, Junjie and Huang, Guan and Zhu, Zheng and Ye, Yun and Du, Dalong},
  journal={arXiv preprint arXiv:2112.11790},
  year={2021}
}

@article{liang2022bevfusion,
  title={Bevfusion: A simple and robust lidar-camera fusion framework},
  author={Liang, Tingting and Xie, Hongwei and Yu, Kaicheng and Xia, Zhongyu and Lin, Zhiwei and Wang, Yongtao and Tang, Tao and Wang, Bing and Tang, Zhi},
  journal={Advances in Neural Information Processing Systems},
  volume={35},
  pages={10421--10434},
  year={2022}
}

@inproceedings{hu2023planning,
  title={Planning-oriented autonomous driving},
  author={Hu, Yihan and Yang, Jiazhi and Chen, Li and Li, Keyu and Sima, Chonghao and Zhu, Xizhou and Chai, Siqi and Du, Senyao and Lin, Tianwei and Wang, Wenhai and others},
  booktitle={Proceedings of the IEEE/CVF conference on computer vision and pattern recognition},
  pages={17853--17862},
  year={2023}
}

@inproceedings{jiang2023vad,
  title={Vad: Vectorized scene representation for efficient autonomous driving},
  author={Jiang, Bo and Chen, Shaoyu and Xu, Qing and Liao, Bencheng and Chen, Jiajie and Zhou, Helong and Zhang, Qian and Liu, Wenyu and Huang, Chang and Wang, Xinggang},
  booktitle={Proceedings of the IEEE/CVF International Conference on Computer Vision},
  pages={8340--8350},
  year={2023}
}

@inproceedings{tong2023scene,
  title={Scene as occupancy},
  author={Tong, Wenwen and Sima, Chonghao and Wang, Tai and Chen, Li and Wu, Silei and Deng, Hanming and Gu, Yi and Lu, Lewei and Luo, Ping and Lin, Dahua and others},
  booktitle={Proceedings of the IEEE/CVF International Conference on Computer Vision},
  pages={8406--8415},
  year={2023}
}

@article{li2024bevformer,
  title={Bevformer: learning bird's-eye-view representation from lidar-camera via spatiotemporal transformers},
  author={Li, Zhiqi and Wang, Wenhai and Li, Hongyang and Xie, Enze and Sima, Chonghao and Lu, Tong and Yu, Qiao and Dai, Jifeng},
  journal={IEEE Transactions on Pattern Analysis and Machine Intelligence},
  year={2024},
  publisher={IEEE}
}

@inproceedings{zhang2022mutr3d,
  title={Mutr3d: A multi-camera tracking framework via 3d-to-2d queries},
  author={Zhang, Tianyuan and Chen, Xuanyao and Wang, Yue and Wang, Yilun and Zhao, Hang},
  booktitle={Proceedings of the IEEE/CVF Conference on Computer Vision and Pattern Recognition},
  pages={4537--4546},
  year={2022}
}

@inproceedings{gu2023vip3d,
  title={Vip3d: End-to-end visual trajectory prediction via 3d agent queries},
  author={Gu, Junru and Hu, Chenxu and Zhang, Tianyuan and Chen, Xuanyao and Wang, Yilun and Wang, Yue and Zhao, Hang},
  booktitle={Proceedings of the IEEE/CVF Conference on Computer Vision and Pattern Recognition},
  pages={5496--5506},
  year={2023}
}

@article{hu2022monocular,
  title={Monocular quasi-dense 3d object tracking},
  author={Hu, Hou-Ning and Yang, Yung-Hsu and Fischer, Tobias and Darrell, Trevor and Yu, Fisher and Sun, Min},
  journal={IEEE Transactions on Pattern Analysis and Machine Intelligence},
  volume={45},
  number={2},
  pages={1992--2008},
  year={2022},
  publisher={IEEE}
}

@article{moon2025mitigating,
  title={Mitigating trade-off: Stream and query-guided aggregation for efficient and effective 3d occupancy prediction},
  author={Moon, Seokha and Baek, Janghyun and Kim, Giseop and Kim, Jinkyu and Choi, Sunwook},
  journal={arXiv preprint arXiv:2503.22087},
  year={2025}
}

@inproceedings{lin2017focal,
  title={Focal loss for dense object detection},
  author={Lin, Tsung-Yi and Goyal, Priya and Girshick, Ross and He, Kaiming and Doll{\'a}r, Piotr},
  booktitle={Proceedings of the IEEE international conference on computer vision},
  pages={2980--2988},
  year={2017}
}

@inproceedings{lu2025toward,
  title={Toward Real-world BEV Perception: Depth Uncertainty Estimation via Gaussian Splatting},
  author={Lu, Shu-Wei and Tsai, Yi-Hsuan and Chen, Yi-Ting},
  booktitle={Proceedings of the Computer Vision and Pattern Recognition Conference},
  pages={17124--17133},
  year={2025}
}

@inproceedings{chabot2025gaussianbev,
  title={Gaussianbev: 3d gaussian representation meets perception models for bev segmentation},
  author={Chabot, Florian and Granger, Nicolas and Lapouge, Guillaume},
  booktitle={2025 IEEE/CVF Winter Conference on Applications of Computer Vision (WACV)},
  pages={2250--2259},
  year={2025},
  organization={IEEE}
}

@inproceedings{huang2025gaussianformer,
  title={Gaussianformer-2: Probabilistic gaussian superposition for efficient 3d occupancy prediction},
  author={Huang, Yuanhui and Thammatadatrakoon, Amonnut and Zheng, Wenzhao and Zhang, Yunpeng and Du, Dalong and Lu, Jiwen},
  booktitle={Proceedings of the Computer Vision and Pattern Recognition Conference},
  pages={27477--27486},
  year={2025}
}

\clearpage

\appendix

\section{Appendix}

The following appendix provides supplementary material related to our proposed GUIDE, including additional details, experiments, discussions, and visualizations that support the main text.


\subsection[\thesubsection]{A. More Details about GUIDE}
\subsubsection{Instance Tracking.}
 Our approach to instance tracking involves calculating unified detection confidence from instance-level features. When this confidence surpasses a predefined threshold, \(T_{\text{track}}\), set at 0.2, the associated instance is linked to the corresponding target and assigned a unique ID. This ID remains consistent throughout temporal propagation, ensuring reliable tracking over time. The simplicity of our tracking method is highlighted by its independence from additional constraints, allowing for efficient and straightforward operation.

\begin{equation}
t_i \leftarrow I_i^{\text{track}}  = s_i^{\text{uni}} \geq T_{\text{track}}
\end{equation}

\subsubsection{Implementation Details.}

In the experiments, we employ GUIDE with input images resized to dimensions of $704\times256$. We adopt ResNet50~\cite{he2016deep} as the backbone network, which is integrated with a Feature Pyramid Network (FPN)~\cite{lin2017feature} to capture multi-scale features at downsampled resolutions of $[1/8, 1/16, 1/32, 1/64]$. Each feature level is configured with a channel dimensionality \( C \) of $256$. And we set the number of instances to $900$, with each instance comprising $48$ Gaussians. All models are trained for up to $100$ epochs utilizing the AdamW optimizer, with the initial learning rate specified as \(2 \times 10^{-4}\). Our experiments are conducted on 8 A100 GPUs with a batch size of $4$ per GPU.

\subsubsection{Comparison with the GaussianFormer Series.}
 

The method in GUIDE, which obtains occupancy via Gaussian representation and Gaussian-to-Voxel splatting, is inspired by the GaussianFormer series of works. GaussianFormer~\cite{huang2024gaussianformer} was the first to pioneer the Gaussian-to-Voxel Splatting method, utilizing 3D Gaussians for 3D semantic occupancy prediction and providing significant computational efficiency gains. Building on this foundation, GaussianFormer-v2~\cite{huang2025gaussianformer} further refined the techniques for computing occupancy probabilities and semantics within Gaussian splatting, effectively reducing the required number of Gaussians. Specifically, the Gaussian-to-Voxel process in GaussianFormer-v2 can be divided into two stages: the first stage infers geometric occupancy results based on the Gaussian probability distribution, while the second stage uses Gaussian opacity and semantic information to obtain the final semantic occupancy classes.

In contrast to the GaussianFormer series, which focuses on semantic occupancy prediction, our GUIDE framework targets instance-level occupancy prediction. The semantic results of occupancy in GUIDE can be directly obtained from the predicted categories of instances. Therefore, we retain only the Gaussian-to-geometric occupancy stage to improve splatting efficiency.


\begin{table}[htp]
\centering
\begin{tabular}{@{}c|c|c|c|c|c@{}}
\hline
\textbf{Methods} & \textbf{mAP$_{0.1}^{8}$↑} & \textbf{mAP$_{0.2}^{8}$↑} & \textbf{mAP$_{0.3}^{8}$↑} \\
\hline
SparseOcc & 21.89 & 13.92 & 7.40\\
\textbf{GUIDE} & \textbf{29.99} & \textbf{21.36} & \textbf{13.47} \\
\hline
\end{tabular}
\caption{\textbf{Details of mAP compared to SparseOcc.} The mAP$^8$ metrics indicates the averge instance occupancy mean Average Precision for 8 foreground  categories, excluding the general obstacle category. SparseOcc and our GUIDE use the same backbone ResNet50 and the same input size 704x256.}\label{tbl6}
\end{table}

\subsection{B. More Experiments}

\subsubsection{Detailed Instance Occupancy mAP Results.}

To provide robust support for our main findings, we conducted comprehensive experimental comparisons of instance occupancy mean Average Precision (mAP) against SparseOcc at varying IoU thresholds \{0.1, 0.2, 0.3\}. As illustrated in \cref{tbl6}, GUIDE consistently surpasses SparseOcc across all three IoU thresholds in evaluating the eight foreground categories. This consistent performance across different thresholds not only underscores the reliability and effectiveness of GUIDE but also reinforces its superior predictive capabilities over SparseOcc. These results highlight the compelling advantages of our approach in accurately determining instance occupancy.





\begin{table}[tbp]
\centering
\begin{tabular}[b]{c|c|cc}
\hline
\makebox[0.14\textwidth][c]{\textbf{Num of Instances}} & \makebox[0.10\textwidth][c] {\textbf{Instance Bank}}  & \makebox[0.06\textwidth][c]{\textbf{mIOU$^{10}$}↑} & \makebox[0.05\textwidth][c]{\textbf{mAP$^{10}_{occ}$}↑} \\
\hline
900 & & {$21.51$} & {$16.42$} \\
900 & \checkmark & \textbf{25.39} & \textbf{22.81} \\
300 & \checkmark & ${22.68}$   & ${18.94}$ \\
\hline
\end{tabular}
\caption{\textbf{Ablation on the number of candidate instances and instance bank utilization.}The metric mIOU$^{10}$ represents the mean Intersection over Union for all 10 foreground categories in the 3D semantic occupancy evaluation. The mAP$_{occ}^{10}$ metric indicates the average instance occupancy mean Average Precision across thresholds $\{0.1, 0.2, 0.3\}$ for 10 foreground categories.}\label{tbl4}
\end{table}

\begin{figure*}[htbp]
    \centering
    \includegraphics[width=0.95\textwidth]{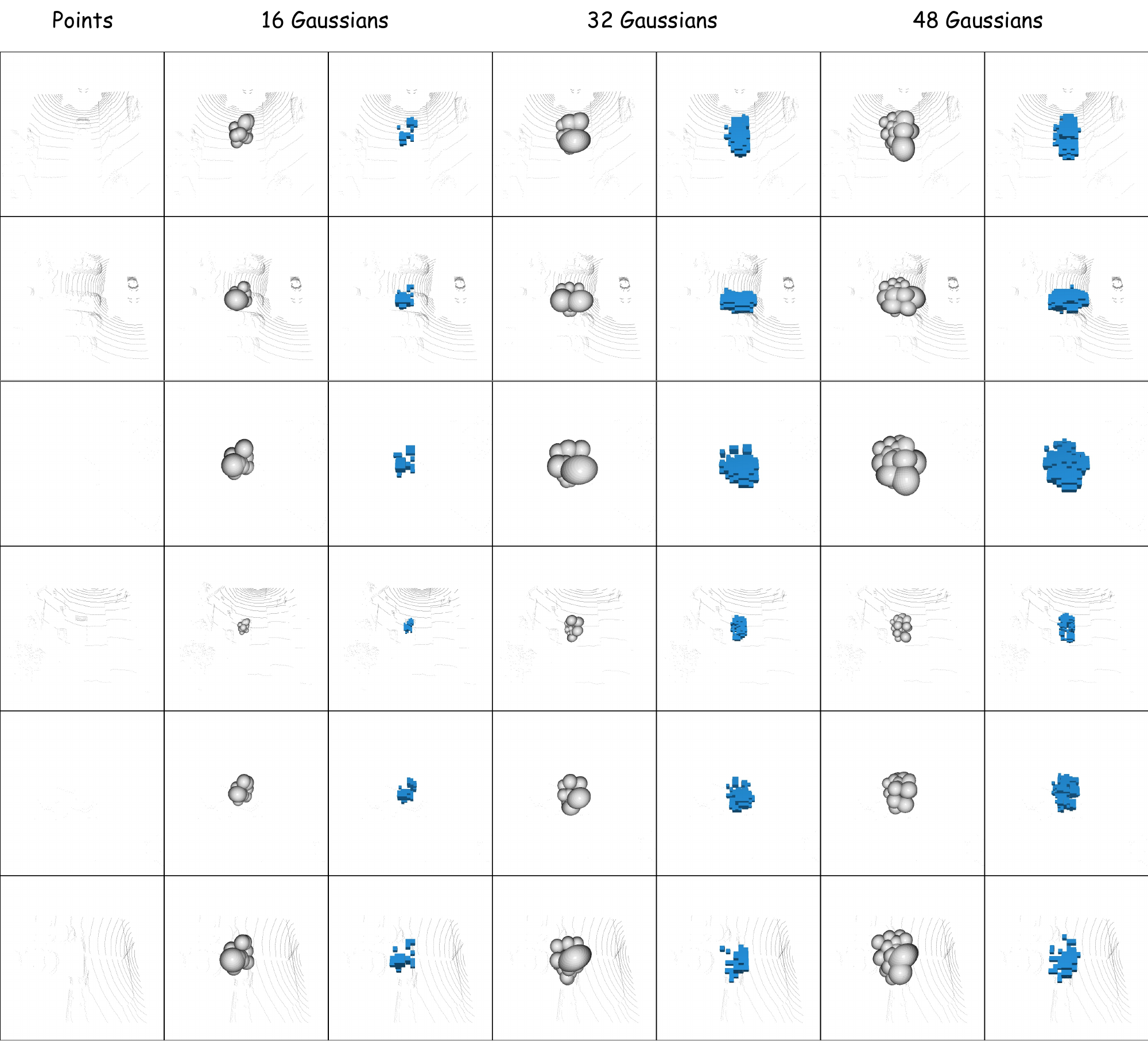}
    \caption{\textbf{Visualization for 3D instance occupancy prediction with different number of Gaussians for each instance.}}
   \label{fig10}
\end{figure*}
\begin{table*}[htbp]
\centering
\begin{tabular}{@{}c|cc|cc@{}}
\hline
\textbf{Number of Gaussians for Each Instance} & \textbf{Latency$\downarrow$} & \textbf{Memory$\downarrow$} & \textbf{mIOU↑} & \textbf{mAP$_{occ}$↑} \\

\hline
16 & \textbf{205 ms} & \textbf{2926 MB} & 22.79 & 18.94 \\
32 & 291 ms & 3154 MB & 23.31 & 20.21\\
48 & 373 ms & 3434 MB & \textbf{25.39} & \textbf{22.81} \\
\hline
\end{tabular}
\caption{\textbf{Details of the ablation study on the number of Gaussians per instance.} The inference latency and memory usage are evaluated using a single NVIDIA 3090 GPU, with a batch size of 1.}\label{tbl7}
\end{table*}

\begin{figure*}[htbp]
    \centering
    \includegraphics[width=0.95\textwidth]{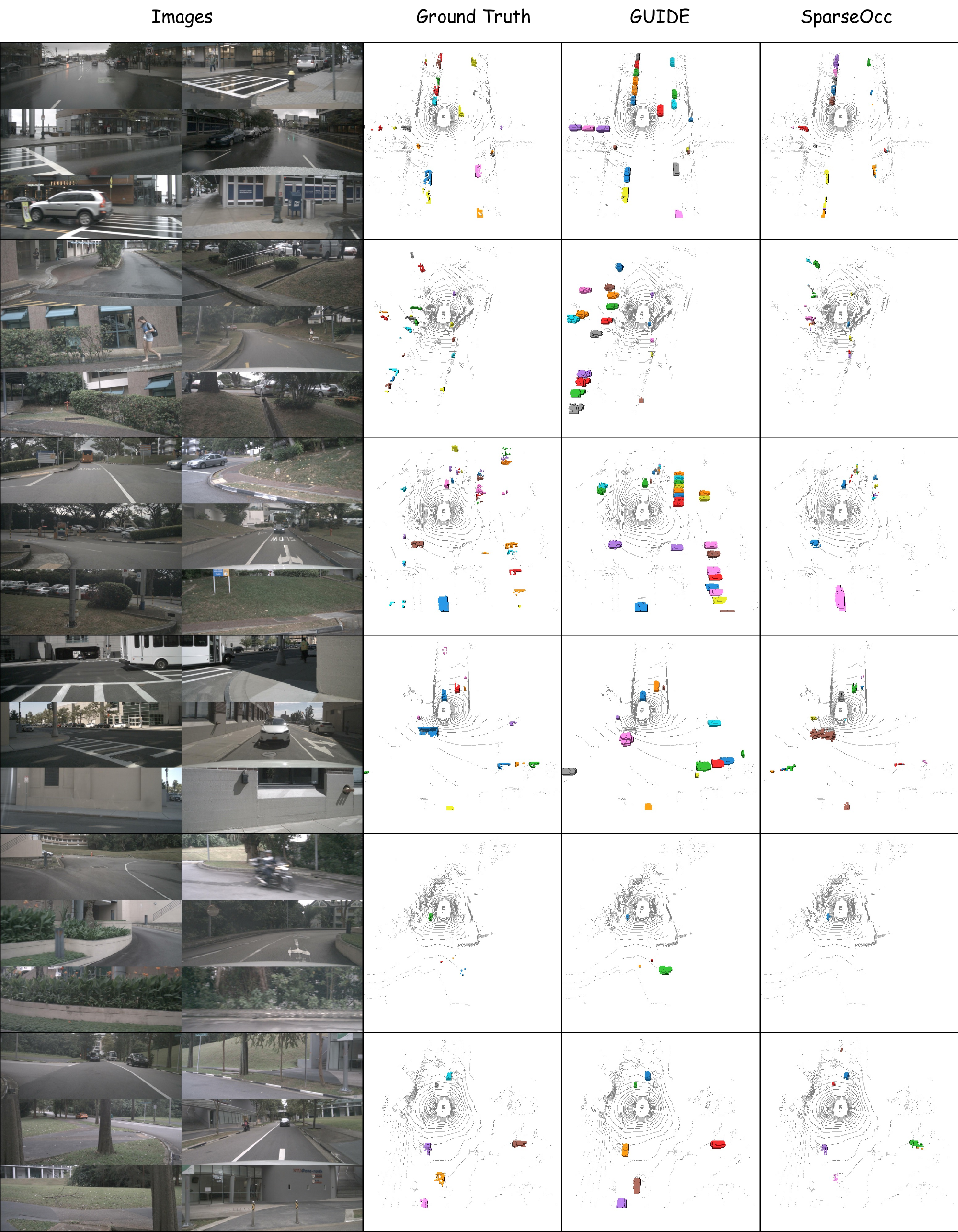}
    \caption{\textbf{More Visualization results for 3D instance occupancy prediction on nuScenes.} The different colors are used to represent the occupancy predictions for various instances. The background points are included to aid scene comprehension and are not utilized during the model inference process.}
   \label{fig1}
\end{figure*}

\subsubsection{Ablations on the Number of Candidate Instances and Instance Bank Utilization.}

We further investigate how the number of candidate instances and the utilization of the instance bank influence the performance of GUIDE. As shown in \cref{tbl4}, increasing the number of candidate instances correlates with improved accuracy in instance occupancy prediction. This improvement may be attributed to the richer scene priors provided by a larger pool of candidate instances, which enhance the model's instance recall and prediction capabilities. 

In terms of the influence of the instance bank, when the number of candidate instances is held constant, utilizing the instance bank also yields improvements in instance occupancy prediction accuracy. This can be attributed to the temporal information supplied by the instance bank, which mitigates the shortcomings of single-frame observations, as well as its tracking ability, which results in more stable instance detection.

\subsubsection{Details of the Ablation Study on the Number of Gaussians per Instance.}

In this section, we explore the effect of using different numbers of Gaussians to model each instance on the performance of our GUIDE. As reported in \cref{tbl7}, our empirical results indicate that increasing the number of Gaussians per instance leads to a corresponding enhancement in the accuracy of instance occupancy predictions. As depicted in \cref{fig10}, this improvement is attributed to the model's increased capacity to represent complex spatial distributions with higher precision.
However, the benefits of using more Gaussians come with increased computational costs. Specifically, as the number of Gaussians grows, there is a noticeable rise in both inference time and GPU memory consumption. This trade-off underlines the critical balance that must be struck between achieving higher prediction accuracy and maintaining computational efficiency.

\subsection{C. More Visualizations}

To further illustrate the superior performance of our GUIDE model in instance recall and occupancy prediction across various scenarios, we present additional visualizations of instance occupancy predictions. As depicted in \cref{fig1}, GUIDE produces significantly denser and more precise instance occupancy predictions when compared to SparseOcc. Moreover, our approach demonstrates enhanced performance in instance recall, as evidenced by aligning the predictions from both methods with the ground truth instances. Notably, SparseOcc has difficulty predicting distant targets, a limitation that GUIDE successfully overcomes. This capability underscores GUIDE’s advantage in overcoming traditional constraints associated with occupancy prediction, enabling an unrestricted prediction range.

\subsection{D. Discussion}

\paragraph{Efficiency.}





GUIDE uses a fully sparse, instance-level Gaussian representation for modeling the occupancy of surrounding obstacles, thus avoiding dependence on dense voxel features. Consequently, the GPU memory usage of GUIDE remains almost unchanged as the perception range increases, providing significant memory efficiency and allowing for larger perception ranges in real-world autonomous driving applications.

\paragraph{Flexible Splatting Resolution.}

Our GUIDE framework employs 3D Gaussians as a dynamic intermediate representation for instance occupancy, enabling flexible adjustment of spatial resolution by modifying voxel size during the Gaussian-to-Voxel Splatting process. This adaptability is particularly advantageous in autonomous driving applications, where perceptual requirements differ depending on an object's proximity to the vehicle. For regions near the vehicle, fine-grained perception is critical for safe decision-making. GUIDE addresses this need by allowing for smaller voxel sizes, thereby providing more precise occupancy predictions for nearby objects. In contrast, for distant objects, increasing the voxel size can significantly reduce computational costs while still maintaining sufficient perception accuracy. This strategy ensures effective monitoring of distant targets without unnecessary resource expenditure. Thus, dynamically adjusting voxel resolution based on proximity optimizes both computational efficiency and perception accuracy. By allowing the voxel size to be varied according to spatial context, GUIDE effectively balances the trade-off between computational cost and prediction accuracy. This flexibility ensures the system adapts to the diverse perceptual demands encountered in real-world driving scenarios, optimizing both performance and resource utilization. Through intelligent voxel size management, GUIDE delivers accurate predictions with reduced computational overhead, meeting the needs of complex and dynamic environments.

\paragraph{Potential for end-to-end Autonomous Driving.}
End-to-end autonomous driving has recently emerged as a research frontier, with modular approaches such as UniAD achieving significant progress. These methods enhance interpretability by transferring instance features from the perception module to the downstream planning module. However, such features are typically derived from bounding box-based detection under auxiliary supervision, resulting in a relatively coarse encoding of structural information. In contrast, our proposed GUIDE framework can be integrated into the perception module to represent detected instances using Gaussian-based structures. This enables the transmission of more detailed spatial information to downstream modules, holding substantial promise for addressing the long-tail distribution of atypical obstacle shapes in autonomous driving scenarios.

\end{document}